\documentclass{article}
\usepackage[utf8]{inputenc}
\usepackage{mathtools,amssymb}
\usepackage{arxiv}
\usepackage{lineno,hyperref}
\usepackage{multirow}
\usepackage{multicol}
\usepackage{booktabs}
\usepackage{enumitem}
\usepackage{colortbl}
\usepackage{tikz}
\usepackage[square,numbers]{natbib}

\usepackage{xcolor}

\title{A Mutation-based Text Generation for Adversarial Machine Learning Applications}

\author{ {Jesus Guerrero} \\
	Department of Computing and Cyber Security\\
	Texas A\&M University - San Antonio\\
	San Antonio, TX 78224 \\
	\texttt{jguer017@jaguar.tamu.eduu} \\
    \And
     {Gongbo Liang}\\
	Department of Computing and Cyber Security\\
	Texas A\&M University - San Antonio\\
	San Antonio, TX 78224 \\
	\texttt{gliang@tamusa.edu} \\
	\And
     {Izzat Alsmadi}\\
	Department of Computing and Cyber Security\\
	Texas A\&M University - San Antonio\\
	San Antonio, TX 78224 \\
	\texttt{ialsmadi@tamusa.edu} \\
}
\hypersetup{
pdftitle={A template for the arxiv style},
pdfsubject={q-bio.NC, q-bio.QM},
pdfauthor={Andrew B.~Trombly},
pdfkeywords={First keyword, Second keyword, More},
}
\begin{document}

\maketitle




\begin{abstract}
Many natural language related applications involve text generation, created by humans or machines. While in many of those applications machines support humans, yet in few others, (e.g. adversarial machine learning, social bots and trolls) machines try to impersonate humans. In this scope, we proposed and evaluated several mutation-based text generation approaches. Unlike machine-based generated text, mutation-based generated text needs human text samples as inputs. We showed examples of mutation operators but this work can be extended in many aspects such as proposing new text-based mutation operators based on the nature of the application.  
\end{abstract}




\section{Introduction}
\label{sec:introduction}
Currently, text generation is widely used in Machine Learning (ML)-based or Artificial Intelligence (AI)-based natural language applications such as language to language translation, document summary, headline or abstract generation. Those applications can be classified into different categories. In one classification, they can be divided into short versus long text generation applications. Short text generation applications include examples such as predicting next word or statement, image caption generation, short language translation, and documents summarization. Long text generation applications include long text story completion, review generation, language translation, poetry generation, and question answering. Large language models such as open AI chat GPT-1,2 and 3 can be used to masquerade humans in many of those short and long text generation applications.
Unlike all previous applications where ML or AI-based text generation exists to support humans and their applications, in online social networks (OSN) such as Twitter and Facebook, text generation is used to fool and influence humans and public opinions through masquerading machine-generated texts as genuinely generated by humans. Social bots and trolls in OSNs are accounts that impersonate humans. They are controlled by ML or AI algorithms. ML algorithms are used to auto generate text in those bot/troll accounts.

As mentioned earlier, ML-based text generation can be used for non-malicious and malicious applications. Our focus in this paper is in malicious applications. There are several malicious ML-based approaches to text generation that can be observed in literature such as:
\begin{itemize}
    \item Classical (e.g. professor/teacher forcing recurrent neural network, RNN): It aims to align generative behavior as closely as possible with teacher-forced behavior, \cite{lamb2016professor}.  
    \item Conventional (e.g. based on hidden Markov models (\cite{jing2002using}), method of moments (MoM), \cite{jones1969field}, Restricted Boltzmann machine (RBM), \cite{holyoak1987parallel} ).
    \item Cooperative training, (e.g. \cite{yin2020meta}, \cite{lu2019cot}).  
    \item Reinforcement based and Reinforcement free approaches (e.g. RankGAN, \cite{lin2017adversarial}, MaliGAN, \cite{che2017maximum}, FMGAN, \cite{chen2018adversarial}, and GSGAN, \cite{kusner2016gans}).
\end{itemize}

In one classification to adversarial machine learning (AML) attacks, those attacks can be divided based on at which machine learning stage the attack is occurring (e.g. (1) learning/training stage, versus (2) testing stage).

\subsection{Poisoning attacks: AML attacks on the learning stage: Manipulating the training data}
Attackers can deliberately influence the training dataset to manipulate the results of a predictive model. A poisoning attack adds poisoned instances to the training set and introduces new errors into the model. 
If we consider one ML application, spam detection, filter of spam messages will be trained with adversary instances to incorrectly classify the spam messages as good messages leading to compromising of system’s integrity. Alternatively, the Spam messages’ classifier will be trained inappropriately to block the genuine messages thereby compromising system’s availability, \cite{newsome2006paragraph}, \cite{perdisci2006misleading}, \cite{nelson2008exploiting}, \cite{rubinstein2009antidote}, \cite{barreno2010security}, \cite{biggio2012poisoning}, \cite{newell2014practicality}. 

\subsection{Evasion attacks: AML attacks on the testing stage: Manipulating the testing data}
\label{sec:Evasion}
In this attack, attackers try to evade the detection system by manipulating the testing data, resulting in a wrong model classification. The core of adversarial evasion attack is that when an attacker can fool a classifier to think that a malicious input is actually benign, they can render a machine learning-based malware detector or intrusion detector ineffective, 
 \cite{russu2016secure}, \cite{zhang2020decision}, \cite{sayghe2020evasion}, \cite{shu2022omni}. \cite{chernikova2022fence}
 
 Our mutation-based approach in this paper can fall under the first category of the last classification (i.e. poisoning attacks). 
A mutation instance is a data instance or object that is created from an original genuine instance with a slight change. Such changes on machine learning stage can reduce accuracy of classification algorithms through generating fake instances that are very close to genuine instances. Mutation changes are typically introduced to simulate actual real-world faults or mistakes. There are several scenarios to implement our mutation-based AML:
\begin{itemize}
    \item Two class labels (Human/Mutation text): Human generated text versus mutation generated text. Such experiments will test classifiers sensitivity to changes injected by mutations in comparison with original genuine text.
    \item Two class labels (Human/Adversarial, mutation instances are added to adversarial instances
    \item Two class labels (Human/Adversarial, mutation instances are added to Human instances
    \item Three class labels (Human/Mutation/Adversarial instances).
\end{itemize}

In this paper, we will propose and evaluate mutation operators within the first category and leave investigating other categories in future papers.

The rest of the paper is organized as follows. Section \ref{sec:Related} provides a summary of related research. Our paper goals and approaches are introduced in \ref{sec:goals}. Section \ref{sec:Experiments} covers the experiments we performed to evaluate our proposed mutation operators. We have then a separate section, section \ref{sec:Comparison} to compare with close contributions. Finally, Section \ref{conclusion} provides some concluding remarks as well as future extensions or directions. 

\section{Related Works}
\label{sec:Related}
Machine learning NLP-based classifiers can be influenced by words misspelling and all forms of adversarial text perturbations.

Literature survey indicated an increasing trend in using pre-trained models in machine learning. Word/sentence embedding models and transformers are examples of those pre-trained models.
Adversarial models may utilize same or similar pre-trained models as well. In another trend related to text generation models, literature showed effort to develop universal text perturbations to be used in both black and white-box attack settings, \cite{alsmadi2022adversarial}.
The literature on adversarial text analysis is quite rich (e.g. see \cite{bravo2020testing}). Our focus is in a selection of those papers on adversarial text generation relevant to poisoning attacks in general or mutation-based approaches in particular.

Word swapping (of semantically equivalent words) attack, \cite{alzantot2018generating} is similar to one example of our mutation operators where one word is swapped from original to mutation instances. \\
In  \cite{alzantot2018generating}, \\ \cite{sayghe2020evasion} and \cite{shi2022word}, words are swapped based on genetic algorithms. Genetic algorithms typically include 5 tasks in which mutation is one of them (Initial population, Fitness function, Selection, Crossover, and Mutation). As alternatives to genetic algorithms, word swapping attacks are also implemented using (1) neural machine translation in , \cite{ribeiro2018semantically}, (2) swarm optimization-based search algorithm, \cite{shi2022word}. In  \cite{alzantot2018generating}, mutation instances are used to fool a sentiment analysis classifier.   

Words substitution can be also role-based lexical substitutions, \cite{iyyer2018adversarial}, or entity-based text perturbation, \cite{liu2022entity}.

In \cite{bhalerao2022data} authors classified intentional and unintentional adversarial text perturbation into ten types, shown in Table \ref{tab-ATB}.

\begin{table}[]

\centering
\scalebox{0.95}{
\begin{tabular}{
>{\columncolor[HTML]{C0C0C0}}l 
>{\columncolor[HTML]{FFFFFF}}l 
>{\columncolor[HTML]{C0C0C0}}l l}
\textbf{Perturbation Type} &
  \textbf{Defense} &
  \textbf{Example} &
  \textbf{Definition} \\ \hline
Combined Unicode &
  ACD &
  \begin{tabular}[c]{@{}l@{}}P.l.e.a.s.e.l.i.k.e.a.n.d.\\ s.h.a.r.e\end{tabular} &
  \begin{tabular}[c]{@{}l@{}}Insert a Unicode character between\\ each original character.\end{tabular} \\
Fake punctuation &
  CW2V &
  \begin{tabular}[c]{@{}l@{}}Pleas.e lik,e abd shar!e the \\ v!deo\end{tabular} &
  \begin{tabular}[c]{@{}l@{}}Randomly add zero or more punc-\\ tuation marks between characters.\end{tabular} \\
Neighboring key &
  CW2V &
  \begin{tabular}[c]{@{}l@{}}Plwase lime and sharr the \\ vvideo\end{tabular} &
  \begin{tabular}[c]{@{}l@{}}Replace character with keyboard-\\ adjacent characters.\end{tabular} \\
Random spaces &
  CW2V &
  \begin{tabular}[c]{@{}l@{}}Pl ease lik e and sha re th e\\ video\end{tabular} &
  \begin{tabular}[c]{@{}l@{}}Randomly insert zero or more \\ spaces between characters.\end{tabular} \\
Replace Unicode &
  UC &
  \begin{tabular}[c]{@{}l@{}}Pleãse lîke and sharê the \\ video\end{tabular} &
  \begin{tabular}[c]{@{}l@{}}Replace characters with Unicode\\ look-alikes\end{tabular} \\
Space separation &
  ACD &
  Please l i k e and s h a r e &
  Place spaces between characters. \\
Tandem character obfuscation &
  UC &
  \begin{tabular}[c]{@{}l@{}}PLE/\textbackslash{}SE LIKE /\textbackslash{}ND \\ SH/\textbackslash{}RE\end{tabular} &
  \begin{tabular}[c]{@{}l@{}}Replace individual characters with\\ characters that together look original\end{tabular} \\
Transposition &
  CW2V &
  Please like adn sahre &
  Swap adjacent characters \\
{\color[HTML]{000000} Vowel repetition and deletion} &
  {\color[HTML]{000000} CW2V} &
  {\color[HTML]{333333} Pls likee nd sharee} &
  Repeat or delete vowels. \\
{\color[HTML]{000000} Zero-width space separation} &
  {\color[HTML]{000000} ACD} &
  {\color[HTML]{333333} \begin{tabular}[c]{@{}l@{}}Please like and share the\\ video\end{tabular}} &
  \begin{tabular}[c]{@{}l@{}}Place zero-width spaces (Unicode\\ character 200c) between characters\end{tabular}
\end{tabular}
}
\caption{Adversarial text perturbations, \cite{bhalerao2022data} }
\label{tab-ATB}
\end{table}


As we mentioned earlier, this work is closely related to AML poisoning attacks,  \cite{newsome2006paragraph}, \cite{perdisci2006misleading}, \cite{nelson2008exploiting}, \cite{rubinstein2009antidote},  \cite{barreno2010security}, \cite{biggio2012poisoning}, \cite{newell2014practicality}.\\
It is also related to AML text perturbations, \cite{vijayaraghavan2019generating}, \cite{eger2019text}, \cite{gao2018black}, and \cite{li2018textbugger}. 

\section{Goals and Approaches}
\label{sec:goals}
According to a previous paper \cite{Wolff2020}, a typical RoBERTa-based classifier mislabels synthetic text to human by very basic differences such as changing 'a's to alpha or 'e's to epsilon. This vulnerability can be used to trick detectors of synthetic text either intentionally or accidentally.
	
To compare synthetic text detectors sensitivity to mistakes or changes to human text we can break up these mutations into operators with the goal of supporting the creation of more generalized synthetic text detectors. Here, these operators will be introduced and be used to fine-tune RoBERTa's pre-trained model to detect mutations, such as the first scenario mentioned earlier, section \ref{sec:Evasion}.

\subsection{Approach: Use a finite set of operators for research customization}
We introduce some examples of mutation operators to implement are in Table \ref{tab-MU-Ops}. These are more advanced can be used for attacking a detector on a more granular level. For our research here however, we will be using more basic mutation operators such as these:

\begin{table}[h!]

\centering
\scalebox{0.95}
{
\begin{tabular}{
>{\columncolor[HTML]{C0C0C0}}l 
>{\columncolor[HTML]{FFFFFF}}l 
>{\columncolor[HTML]{C0C0C0}}l }
\textbf{Mutation Operator} &
  \textbf{Example} &
  \textbf{Definition} \\ \hline
Randomization &
  \begin{tabular}[c]{@{}l@{}}Plz sh\ensuremath{\alpha}r and hate \\ film\end{tabular} &
  \begin{tabular}[c]{@{}l@{}}Use all below mutation \\ operators\end{tabular} \\
Misspelling words &
  \begin{tabular}[c]{@{}l@{}}Plz sharr and like the \\ vid\end{tabular} &
  Misspell a few words \\
Deleting articles &
  \begin{tabular}[c]{@{}l@{}}Please share and like\\ video\end{tabular} &
  \begin{tabular}[c]{@{}l@{}}Delete a few articles, \\ including starting ones\end{tabular} \\
\begin{tabular}[c]{@{}l@{}}Random word with \\ random word\end{tabular} &
  \begin{tabular}[c]{@{}l@{}}Please roar and tree \\ video\end{tabular} &
  \begin{tabular}[c]{@{}l@{}}Replace a random word\\ with another random word\end{tabular} \\
Synonym replacement &
  \begin{tabular}[c]{@{}l@{}}Please disseminate and\\ prefer the video\end{tabular} &
  \begin{tabular}[c]{@{}l@{}}Replace a word with its\\ synonym\end{tabular} \\
Antonym replacement &
  \begin{tabular}[c]{@{}l@{}}Please hide and hate \\ the video\end{tabular} &
  \begin{tabular}[c]{@{}l@{}}Replace a word with its\\ antonym\end{tabular} \\
Replace "a", "e" &
  \begin{tabular}[c]{@{}l@{}}Please like  \ensuremath{\alpha}nd share \\ the video\end{tabular} &
  \begin{tabular}[c]{@{}l@{}}Replace some a's and e's\\ with epsilon \& alpha\end{tabular}
\end{tabular}
}
\caption{Experimental mutation operators}
\label{tab-MU-Ops}
\end{table}

These 7 operators can replicate simple mistakes and changes which can happen to human written text, including the 2 operators used in previous cited works. These were chosen for their ease of implementation and usage in previous research. We will leave more advanced and numerous operators for future papers.

As for the implementation, most of these operators use word maps which iterate through each string replacing the words with the intended character, word insertion or deletion. Punctuation and excess special characters are removed for simplicity. Though there is different methods for the different mutation operators.

The random word operator is in fact a list of random words. Arbitrary words are chosen and replaced with a random word from the same list. Limits to the number of mutations should be added to limit the operator from completely fuzzing the string as well. The code from our GitHub, \url{https://github.com/JesseGuerrero/Mutation-Based-Text-Detection} has some written operators which can be viewed as examples.

In our implementation word maps are limited to 3000 of the most common words, synonyms and antonyms. These words can be pulled from any API service such as RapidAPI to get lists of words, synonyms, adverbs, verbs, etc.

\subsection{Approach: Test mutations by Evasion attacks on neural network detectors}
We will be testing these 7 operators against RoBERTa pre-trained models. Of these 7 operators we want to test how they will affect a previously researched synthetic text detector, how a fine-tuned mutation text detector will be affected and as well as the differences between the different mutation operators. Lastly we want to see how shorter text affects these results. More details on the next section, \ref{sec:Experiments}.

\section{Experiments and Analysis}
\label{sec:Experiments}
With these mutations, we can introduce mutation detection with a classic binary RoBERTa classifier. This part of the experiment is the extension portion of the previous author's work mentioned before with an actual solution to the vulnerabilities in that paper.

\subsection{Experiment methodology}
We used an existing RoBERTa classifier which is meant to classify synthetic and human generated text to test how it would classify mutated text. It is still the pre-trained binary model, however it is being retrained to detect mutation rather than synthetic text.

The data set used was the full COCO images data set where hand written captions are placed for each image. A total of 5 captions are human created per image. The captions were parsed into a re-usable format and were used to train the human portions of the model.

Across the training, testing and validation sets there were over 700,000 human texts used to train the model. Two models were made from this data set. The first was based on individual captions. The second was based on these 5 captions combined per unique image name for calling via a map.  

Six operators were used as mutations for this classifier. They are; (1) replacing synonyms, (2) antonyms, (3) random words, (4) removing articles, (5) replacing a with alpha and e with epsilon, and lastly, (6) the most common misspellings.

The training data was duplicated for the mutation data sets. Over 700,000 texts were used for training and the same texts were re-used for mutations. This meant for individual captions there were over 1.4 million text instances with both mutation and human labels. For combined captions there were 1/5 of the total instances.

The data sets were selected as they were already labelled from COCO dataset. The training set went to training, the validation set went to validation and testing to testing. For training the mutation label, the mutation operators were used at run-time. 

The operator was randomly chosen at run-time with a simple random function among 6 operators. Each operator was used so the classifier can learn to detect all 6 of these mutation types in one classifier.

So far as testing is concerned, the same testing data set from COCO was re-used with an operator manipulating a whole set. A total of 7 testing data sets were created for each of the 6 operators and a seventh randomized data set, like the mutations at run-time. This formed our metrics of how accurately the model can correctly label each instance of the mutation data sets as mutations and how accurately the model can label human text. In a total 8 operators, 1 human and a seventh randomizing the first six mutations.

\subsection{Preliminary Results}
If we were to apply these operators to the previously researched synthetic text we should get poor results for detecting mutated text. Given text derived from human text, though just modified, is still synthetic, we can see that mutation poses a vulnerability to detecting machine and human generated texts. Here are the results: 

\begin{table}[h!]
\centering
\begin{tabular}{
>{\columncolor[HTML]{C0C0C0}}l 
>{\columncolor[HTML]{FFFFFF}}l }
\textbf{Operator Type} & \cellcolor[HTML]{FFFFFF}\textbf{Accuracy}           \\ \hline
None                   & \cellcolor[HTML]{FFFFFF}$\sim$88.80\%(1000 samples) \\
Randomized             & $\sim$01.00\%(1000 samples)                         \\
Replace Alpha, Epsilon & $\sim$01.01\%(1000 samples)                         \\
Misspelling words      & $\sim$00.00\%(1000 samples)                         \\
Delete articles        & $\sim$01.60\%(1000 samples)                         \\
Synonym replacement    & $\sim$00.00\%(1000 samples)                         \\
Replace random word    & $\sim$07.79\%(1000 samples)                         \\
Antonym replacement    & $\sim$09.89\%(1000 samples)                        
\end{tabular}
\caption{Preliminary Results}
\end{table}
As we can see from 1000 samples the accuracy is quite poor when modifying the text. The original detector without mutations had a recall of over 97\% detection of synthetic and human text in-distribution. For our research outside of the paper we have an out of distribution pure human text data set as 88\% accurate as the 1st row in the table. 

This means the detector is quite good at classifying human text out of distribution and is even good at detecting in-distribution synthetic text. The model does those things above human distinction which in the past was around 54\% accurate. Our issue from our modeling is mutation from which the model does not perform well.

\subsection{Experiments results \& analysis}
So far as the first run through with individual captions, the results were pretty good. A total of 2,490 texts were tested for the detector from the testing data set. In total overall the detector accuracy was about 91\% and each epoch took 13 hours for a total of 4 epochs or 52 hours total.

\begin{table}[h!]
\centering
\begin{tabular}{
>{\columncolor[HTML]{C0C0C0}}l 
>{\columncolor[HTML]{EFEFEF}}l }
\textbf{Operator Type} & \textbf{Accuracy}   \\ \hline
None                   & $\sim$71.48\%(2490) \\
Randomized             & $\sim$99.83\%(2490) \\
Replace alpha, epsilon & $\sim$99.95\%(2490) \\
Misspelling words      & $\sim$99.95\%(2490) \\
Delete articles        & $\sim$59.87\%(2490) \\
Synonym replacement    & $\sim$99.91\%(2490) \\
Replace random word    & $\sim$100\%(2490)   \\
Antonym replacement    & $\sim$99.03\%(2490)
\end{tabular}
\caption{Individual captions, short language modeling}
\end{table}

The most inaccurate operator overall was always the "delete articles operator". This can be due to some semantic issues or just the difficulty of detecting what is \emph{not} there rather than what \emph{is there} to a RoBERTa classifier. For this first run, the other operators ranged from 59\% accurate to 100\% accurate, with human detection being 71\%.

For the second run the captions for text chunks were combined per image. This meant all 5 captions were now one text and were fed to the training model. This means 1/5 th of the instances but more per text. The results were a slightly lower; total overall detector accuracy of ~88\% with 2490 texts being tested.

The epochs took about 2 hours each this way as well. A total of 10 epochs or 20 hours were used to finish the model training. Same as before, the delete articles operator was the weakest mutator and was in fact even weaker the second time.

\begin{table}[h!]
\centering
\begin{tabular}{
>{\columncolor[HTML]{C0C0C0}}l 
>{\columncolor[HTML]{EFEFEF}}l
}
\textbf{Operator Type} & \textbf{Accuracy}   \\ \hline
None                   & $\sim$93.65\%(2490) \\
Randomized             & $\sim$98.96\%(2490) \\
Replace alpha, epsilon & $\sim$99.92\%(2490) \\
Misspelling words      & $\sim$99.80\%(2490) \\
Delete articles        & $\sim$25.42\%(2490) \\
Synonym replacement    & $\sim$99.76\%(2490) \\
Replace random word    & $\sim$98.43\%(2490) \\
Antonym replacement    & $\sim$92.37\%(2490)
\end{tabular}
\caption{Combined captions, longer language modeling}\label{fig3}
\end{table}

Besides the weakest operators, the other operators were all above 95\%, much better than the original neural network. If we were to remove the lowest performing operator we would in fact have 95\% accuracy for the 1st run and 97\% accuracy for the 2nd. This means in reality the combined text may be the better approach to train. 

\subsection{Experiment: Models issues and weaknesses}
The main issue of the evaluated models is possible a bias issue. It seems that models work mostly for semi in-distribution data sets. So accuracy is altered quite a bit by outside data sets. Both the individually captioned model and the grouped model were tested in an out of distribution data set. The individually captioned data set was 2.2\% accurate for human detection and 100\% accurate for Alpha/Epsilon mutation, while the grouped captioned data set was 55.7\% accurate for human detection and 100\% for Alpha/Epsilon mutation. The issue appears to be the out of distribution set may have had vocabulary that didn't exist in distribution and the detector defaults to the mutation label. Still in this sense, the longer the text-set the better is the performance.  

In the future we can use this work flow with more diverse training data sets to generalize models to out of distribution and use these models to prevent simple mutation from fooling binary detectors. Lastly, the mutation operator "delete articles" seems like a great way to fool the classifier into mislabeling mutated text. 

\section{A Comparison Study} 
\label{sec:Comparison}
\subsection{Machine Text Generation}

Machine text generation is a field of study in Natural Language Processing (NLP) that combines computational linguistics and artificial intelligence that has progressed significantly in recent years. Neural network approaches are widely used for this task and keep dominant in the field. The state-of-the-art methods may include Transformers ~\cite{vaswani2017attention}, BERT~\cite{devlin2018bert}, GPT-3~\cite{brown2020language}, RoBERTa~\cite{liu2019roberta}, etc. 
The models are trained on a large amount of text data. For example, the GPT-2 model was trained on text scrapped from eight million web pages~\cite{radford2019language}, and is able to generate human-like text. Due to the high text generation performance, such methods are very popular on tasks, such as image caption generation, text summarization, machine translation, moving script-writing, and poetry composition. However, the output of such methods is often open-ended. 


Through this work, we propose a mutation-based text generation method that can be distinguished from the existing text generation method fundamentally. Unlike the neural network based methods, the mutation-based method generates output based on the given text under a given condition. The text is generated in a tightly controlled environment, and the output is closed-ended. The well-controlled environment makes the output of the mutation-based method suitable for serious security test tasks, such as machine learning model vulnerability tests and SQL injection defense and detection.

Given a text corpus (e.g., a sentence or a paragraph), $\mathcal{T}$, which contains an ordered set of words, $\mathcal{W}=\{w_1, w_2, ..., w_n\}$, and an ordered set of punctuation, $\mathcal{P}=\{p_1, p_2, ..., p_m\}$, a mutation operator, $\mu(\cdot)$ is used to generate the mutation-based text. For instance, given a character-level mutation operator, $\mu_c(\cdot)$:
\begin{equation}
    \mathcal{W}' = \mu_c(w_i, \rho, \sigma),
\end{equation}
where $\mathcal{W}'$ is the output of $\mu_c(\cdot)$, which replaces the letter $\rho$ in $w_i$ ($w_i \in \mathcal{W}$) with a mutation $\sigma$. Then, the final output of $\mathcal{T}$ is $\mathcal{T}'=\langle\mathcal{W}', \mathcal{P}\rangle$. For instance, assume $\mathcal{T}=$ \texttt{"Text generation is interesting!"}, $\mathcal{W} = \{ \texttt{Text}, \texttt{generation}, \texttt{is}, \texttt{interesting}\}$, and $\mathcal{P} = \{\texttt{!}\}$, $w_i = \texttt{generation}$, and a character-level mutation operator $\mu_u(\cdot)$, where $\rho=\texttt{a}$ and $\sigma=\alpha$ (the Greek letter alpha). Then, the output text corpus, $\mathcal{T}'$ is generated as:
\begin{equation}
    \begin{split}
    \mathcal{T}' & = \langle\mathcal{W}', \mathcal{P}\rangle \\
    & = \langle\mu_c(w_i, \rho, \sigma), \mathcal{P}\rangle \\
    & = \langle\mu_c(\texttt{generation}, \texttt{a}, \alpha), \mathcal{P}\rangle \\
    & = \texttt{Text gener}\alpha\texttt{tion is interesting!}
    \end{split}
\end{equation}

\subsection{Mutation Testing in the Language Domain}

The proposed mutation-based text generation is
inspired by the advances in mutation analysis in software testing and the idea of "broiling frog syndrome" The well-controlled and close-ended environment makes generating precise text output possible. By using the proposed tool, researchers could test a language model by changing the input slightly and step-by-step. 

Charm~\cite{bravo2020testing} is a chat-bot testing tool closely related to our mutation-based text generator that extended Botium~\cite{jin2020bert}---a popular framework for chat-bot testing---by integrating eight mutation operators. Though both Charm and ours use mutation operators, Charm is proposed as an extension to Botium and relies on Botium to work. In addition, Charm only works for chat-bot testing. 

Unlike Charm, ours is a general-purpose language generation method, which is designed to work alone. Our method can be used to analyze any type of language models that accept a sequence of text as input. In addition, the users are not limited by the pre-defined mutation operators. Our proposed mutation-based text generator is a general framework. The users could easily design their own mutation operators for their specific tasks within our framework.

\section{Discussions, Conclusions and Future Directions}
\label{conclusion}
Automatic text generation techniques are adopted into various domains, from question-answering to AI-driven education. Due to the progress of neural network (NN) techniques, NN-based approaches dominate the field. Though advanced techniques may be applied to control text generation direction, the text is still generated in a widely open-ended fashion. For instance, a NN-based approach can generate a greeting message to greet a specific person. However, it is hard to control the exact wording used in the message. Such open-ended text generation might work fine for content generation tasks. However, due to lack of precise control, using open-ended text generation methods to systematically evaluate flaws in language analysis models may be non-trivial.

Unlike the existing language models, our proposed mutation-based text-generation framework provides a tightly controlled environment for text generation that extends text-generation techniques to the field of cyber security (i.e., flaws evaluation for language analysis models). The output of our framework can be used to systematically evaluate any machine learning models or software systems that use a sequence of text as input, such as SQL injection detection~\cite{hlaing2020detection} and software debugging~\cite{zhao2022recdroid}. Researchers may also design their own mutation operators under our framework.

We demonstrated the proposed text-generation framework using the RoBERTa-based detector that is pre-trained for separating human-written text from synthetic ones. Our experiments showed that the RoBERTa-based detector has a significant flaw. As a detection method, it is extremely vulnerable to simple adversarial attacks, such as replacing the English letter "a" with the Greek letter "$\alpha$" or removing the articles---a, an, the---from a sentence. We also demonstrated that simply including the adversarial samples (i.e., the mutation texts) in the fine-tuning stage of the classifier would significantly improve the model robustness on such types of attack. However, we believe that this issue should be better addressed on the feature level since any changes at the text level will lead to changes in the tokenization stage that will eventually lead to a different embedding vector being fed into the classification network. Thus, one future direction of this work is reducing the distance between the original and mutation samples in the feature space. Some potentially useful methods might include using  contrastive learning and siamese network~\cite{koch2015siamese,liang2021contrastive} as well as dynamic feature alignment~\cite{zhang2021dynamic,dong2020cscl}.

Besides improving the robustness of the RoBERTa-based detector, we plan to continue to work on the development of the proposed mutation-based text generation framework. Currently, the framework only works in a two-step testing scenario. Users need to use the framework to generate the testing cases and feed the testing cases into the downstream model in separate steps. We plan to release a library that can be directly imported into any downstream applications. Tools for easily creating and editing mutation operators will also be created. In addition, a graphical user interface may also be developed.

In conclusion, we propose a general-purpose, mutation-based text generation framework that produces close-ended, precise text. The output of our framework can be used in various downstream applications that take text sequences as input, providing a systematic way to evaluate the robustness of such models. We believe the proposed framework offers a new direction to systematically evaluate language models that will be very useful to those who are seeking insightful analysis of such models.

\bibliography{MutationAMLArXiv}  

\end{document}